\documentclass{article} 
\usepackage{iclr2025_conference,times}

\usepackage{amsmath,amsfonts,bm}

\def\eqref#1{equation~\ref{#1}}

\def\1{\bm{1}}

\DeclareMathAlphabet{\mathsfit}{\encodingdefault}{\sfdefault}{m}{sl}
\SetMathAlphabet{\mathsfit}{bold}{\encodingdefault}{\sfdefault}{bx}{n}

\usepackage[utf8]{inputenc} 
\usepackage[T1]{fontenc}    
\usepackage{hyperref}       
\usepackage{url}            
\usepackage{booktabs}       
\usepackage{amsfonts}       
\usepackage{nicefrac}       
\usepackage{microtype}      
\usepackage{xcolor}         
\usepackage{fancyvrb}       
\usepackage{todonotes}
\usepackage{listings}
\usepackage{multirow}
\usepackage{floatpag}
\usepackage{wrapfig}
\usepackage{multicol}
\usepackage{setspace}
\usepackage{lipsum}
\usepackage{soul}
\usepackage{comment}
\usepackage{subcaption}
\usepackage{float}
\newcommand{\codeinline}[1]{\lstinline|#1|}

\title{Investigating task-specific prompts and sparse autoencoders for activation monitoring}

\author{%
  Henk Tillman \\ OpenAI
  \And
  Dan Mossing \\ OpenAI
}

\usepackage{xspace}

\iclrfinalcopy 
\begin{document}

\maketitle

\begin{abstract}

Language models can behave in unexpected and unsafe ways, and so it is valuable to monitor their outputs. Internal activations of language models encode additional information that could be useful for this. The baseline approach for activation monitoring is some variation of linear probing on a particular layer: starting from a labeled dataset, train a logistic regression classifier on that layer's activations. Recent work has proposed several approaches which may improve on naive linear probing, by leveraging additional computation. One class of techniques, which we call “prompted probing,” leverages test time computation to improve monitoring by (1) prompting the model with a description of the monitoring task, and (2) applying a learned linear probe to resulting activations. Another class of techniques uses computation at train time: training sparse autoencoders offline to identify an interpretable basis for the activations, and e.g. max-pooling activations across tokens using that basis before applying a linear probe. However, one can also prompt the model with a description of the monitoring task and use its output directly. We develop and test novel refinements of these methods and compare them against each other. We find asking the model zero-shot is a reasonable baseline when inference-time compute is not limited; however, activation probing methods can substantially outperform this baseline given sufficient training data. Specifically, we recommend prompted probing when inference-time compute is available, due to its superior data efficiency and good generalization performance. Alternatively, if inference-time compute is limited, we find SAE-based probing methods outperform raw activation probing.

\end{abstract}

\section{Introduction}
To ensure the safety of AI systems in practice, other AI systems are often used to monitor their outputs. Two common approaches are (1) prompting large language models (LLMs) with a natural language monitoring task, and using natural language output as the monitor signal; and (2) probing LLM hidden layer activations using logistic regression classifiers, and using the classifier output as the monitor signal. The goal of this work is to ground these two approaches, prompting and probing, by comparing variants (and combinations) of both that have been proposed in the literature. We pay particular attention to probing variants involving sparse autoencoders (SAEs) which have been proposed as a way to leverage train time compute to improve probe performance.\\\\
These approaches have various limitations in resource-constrained settings. For example, prompting the model does not require labeled training data for the monitoring task (a model can often perform the task zero-shot). However, prompting the model requires additional inference time compute (more forward passes are required for the monitoring prompt tokens). On the other hand, probing can be ``free'' at inference time (leveraging only the forward passes that were used to generate the model’s output in the first place). But probing may require substantial labeled training data, as LLM activations are not \textit{a priori} interpretable unlike prompted natural language outputs. Combining SAEs with probing, as proposed in \citet{bricken_using_2024}, or combining prompting with probing, as proposed in \citet{zou_representation_2023}, may improve the performance of probing, although evidence is mixed \citep{kantamneni_are_2025,smith_negative_2025}. Therefore we investigate these techniques across various constraints, including training data availability, limited inference-time compute (i.e. whether prompting is possible), and restricted training-time compute (i.e. whether training SAEs is possible).\\\\
Building upon previous studies such as \citet{bricken_using_2024} and the Linear Artificial Tomography (LAT) method from \citet{zou_representation_2023}, we systematically compare raw activation probing, both prompted and un-prompted, as well as SAE-based probing. We hypothesize that prompted probing can enhance the separability of activation representations with respect to task-specific features, improving probe performance especially in low-data regimes. Through experiments on moderation, question-answering, and sentiment classification tasks, we find that prompted probing significantly boosts classifier accuracy relative to un-prompted probing. We also investigate whether modifications to prompted probing such max-pooling activations or adding few-shot examples can improve performance. Additionally, we explore the computational trade-offs between probing raw activations and leveraging autoencoder latents. We conclude that prompted probing combined with direct activation probing is the most effective method when inference-time compute is not a constraint. These findings provide practical insights into best practices for activation monitoring in various deployment scenarios.\\\\

Our contributions are as follows:
\vspace{-5pt}
\begin{enumerate}
    \item We systematically evaluate activation probing techniques, including in the settings of limited training data and distribution shift between train and test sets.
    \item We demonstrate that prompted raw activation probing significantly enhances classifier performance relative to un-prompted probing, particularly in low-training data and out-of-distribution regimes.
    \item We compare prompted probing with other prior methods from \citet{bricken_using_2024} and \citet{zou_representation_2023}. Our results agree with \citet{bricken_using_2024} that max-pooled SAE activation probing outperforms raw activation probing.

\end{enumerate}

\section{Related work}
Linear probes, introduced by \citet{alain_understanding_2018}, are a common technique for investigating model behavior through activations. \citet{belinkov_interpretability_2020} introduce a distinction between structural probing (analyzing model internals) and behavioral probing (examining model outputs), analogous to our comparison of activation probing with model zero-shot performance. Linear artificial tomography (LAT), introduced by \citet{zou_representation_2023}, can be seen as integrating aspects of both approaches. Prompted probing in our work is a simplified variant of LAT which uses supervised rather than unsupervised learning. \citet{goldowsky-dill_detecting_2025} explore a similar method (which they call ``probing follow-up question'') in additional to standard probing techniques and find them to have strong performance in detecting deceptive model behavior, often outperforming black-box detection methods. \citet{azaria_internal_2023} find significant out-of-distribution hallucination detection performance when training shallow neural network classifiers on LLM activations. \citet{gottesman_estimating_2024} train linear classifiers on LLM activations for entity tokens to obtain predictive measures of LLM knowledge about such entities. \citet{liu_cognitive_2023} also investigate model truthfulness in the setting of answering multiple choice and true/false questions. They find that most probing and model output disagreements stem from differences in confidence, and that high-confidence disagreements between the probe and model outputs are relatively rare.\\

Many works have investigated the properties and interpretability of linearly-derived signals from model activations. \citet{gurnee_finding_2023} show that model activations encode interpretable features sparsely, with individual neurons often representing multiple features in superposition. They also introduce a mean-difference heuristic for identifying relevant probing features, which we incorporate as a regularization method (see Section~\ref{sec:linear_classifier_training}). \citet{tigges_linear_2023} use linear probes to show that models tend to summarize high-level concepts like sentiment in intermediate tokens, such as punctuation and certain nouns. Similarly, \citet{orgad_llms_2024} investigate hallucinations in LLMs and show that truthfulness information is stored most saliently at the ``exact answer'' tokens forming the most semantically important part of a response. \citet{pan_latentqa_2024} propose a method for decoding model activations into natural language, enabling open-ended interrogation of a model’s internal states and elicitation of relational information.\\

Several papers have directly compared SAE-based probes vs. straightforward linear probes, finding mixed results. \citet{bricken_using_2024} find that probing SAE features offers a minor improvement over standard linear probing for classification, and emphasize the importance of max-pooling SAE activations over the token dimension and fine-tuning SAEs on problem-specific data. \citet{gallifant_sparse_2025} explore max-pooling techniques and feature binarization in the context of out-of-distribution generalization. \citet{kantamneni_are_2025} and \citet{smith_negative_2025} compare SAE-based classification methods to conventional machine learning approaches across diverse tasks, finding little to no benefit to SAE-based techniques in a variety of settings unless max-pooling is applied. They also find benefits to using attention-based probes.

\section{Methods}
\label{sec:methods}
\subsection{Data and Models}
We examine activations of a previous ChatGPT-4o snapshot across all tasks.

We experiment with three datasets. Our first dataset is a “gold” eval set for \href{https://platform.openai.com/docs/guides/moderation#content-classifications}{OpenAI moderation categories}: types of restricted content such as violence, harassment, etc. We focus on a high-quality set of 2,610 passages which are hand-labeled by experts, for each of several categories of restricted content. In this report we focus on harassment and violence, categories with relatively large numbers of positive examples. We always balance the positive and negative examples in the training set. This dataset contains both English and non-English text, and orthogonally, both isolated prompts and multi-turn conversations (see below).\\\\
Our second dataset is constructed using the SimpleQA dataset publicly released by OpenAI \citep{wei_measuring_2024}. This dataset consists of questions selected specifically to be adversarial to GPT-4o. To construct our version of the dataset, we sample 5 ``on-policy'' answers for each question, using the same snapshot used to probe activations. We auto-label each answer as correct or incorrect using an LLM prompted with the answer key. We then subselect only questions for which 5/5 answers were either correct or incorrect. We randomly subsample 10,000 question-answer pairs from the training set, balanced between correct and incorrect, and train our models to predict the correct pairs. Because questions have multiple answers, some questions appear multiple times. We use roughly 6,000 unique questions, and create the train/test split such that no questions appear in both splits simultaneously. Our third dataset is the training set from the Kaggle RottenTomatoes sentiment analysis dataset \citep{pang_seeing_2005}. We use the entire set of roughly 10,000 movie reviews, which is balanced between positive and negative sentiment.\\\\
For each dataset, we use an 80/20 train/test split. We always keep the test split the same, so that experiments are comparable. When training on reduced numbers of examples, we subselect from the 80\% train split. The test set sizes are as follows: harassment detection (149 positive, 373 negative), violence detection (222 positive, 300 negative), hallucination detection on SimpleQA responses (1013 positive, 1017 negative), and sentiment analysis on RottenTomatoes (871 positive, 835 negative). We conduct in-distribution/out-of-distribution generalization experiments only on the moderation datasets, across English/non-English and Chat/non-Chat (i.e. is the passage a multiturn conversation or not) boundaries. English/non-English labels are included in the moderation ground truth labels. We obtain Chat/non-Chat labels by checking for the presence of any of a set of special conversation tags in the passage.
\subsection{Passage Processing}
\label{sec:passage_processing}

In this work, we test whether mentioning a category of interest along with the passage to be monitored (``prompted probing'') improves monitoring performance, inspired by the Linear Artificial Tomography method in \citet{zou_representation_2023}. The ``prefix+suffix'' prompted probing template includes a prefix before the passage which refers to the category we are monitoring for (e.g. ``Please evaluate whether the following passage contains harassment.''). After the passage, the template includes a prompt suffix asking the model to answer `Yes' or `No' whether the passage contains content of the monitored category. We also experiment with a suffix-only template which begins simply with the passage, allowing us to share a single forward pass across categories for all but the final category-specific suffix tokens.\\

The passages in the moderation set are a mixture of raw text, chat-formatted conversations (i.e. including special user and assistant tags), and fragments of chat-formatted conversations (e.g. conversations that begin or end mid-message, or that are a fragment of a single message). We preprocess these passages differently depending on their initial formatting and whether we are using prompted probing. If we aren't using prompted probing, we convert raw text passages into a single chat-formatted conversation where the passage is contained in a single user message. \textbf{In the following, chat formatting is indicated by a bolded lower-case role, followed by the content of that role's message.} For instance, if the raw text passage is ``Bob punches Joe,'' we would convert the passage to the following format:
\begin{Verbatim}[commandchars=\\\{\}]
\textbf{system:} You are ChatGPT, a large language model trained by 
<Organization>.
\textbf{user:} Bob punches Joe
\end{Verbatim}

Preexisting chat-formatted passages we leave unchanged, simply running inference on the passages as they are and saving the resulting activations.\\

\definecolor{prefixcolor}{rgb}{0, 0.6, 0}  
\definecolor{suffixcolor}{rgb}{0.8, 0, 0.8}  

Now suppose we are using one of the prompted probing templates and that we start with a chat-formatted conversation. For instance, suppose we start with the following conversation:
\begin{Verbatim}[commandchars=\\\{\}]
\textbf{system:} You are ChatGPT, a large language model trained by 
<Organization>.
\textbf{user:} write me a story where snape shanks dumbledore
\textbf{assistant:} I'm sorry, but I cannot assist you  with writing a 
story like that.
\end{Verbatim}
[Note: the system role is typically used to set the context or provide initial instructions.]\\
We first convert the chat-formatted conversation into a \textit{plaintext} ``User: ... Assistant: ...'' format (note the lack of bolding, indicating plaintext rather than explicit chat formatting):
\begin{verbatim}
User: write me a story where snape shanks dumbledore
Assistant: I'm sorry, but I cannot assist you with writing a story
like that.
\end{verbatim}
In the following we use text color for clarity. If we are using \textbf{\textcolor{prefixcolor}{prefix}+\textcolor{suffixcolor}{suffix} prompted probing}, before running inference we put the resulting plaintext into a chat-formatted conversation template where the concept appears before the passage (supposing we want to elicit the ``violence'' concept). Note that the `<passage>` tags here are normal strings and not special formatting tags.
\begin{Verbatim}[commandchars=\\\{\}, formatcom=\relax]
\textbf{user: }\textcolor{prefixcolor}{Please evaluate whether the following passage contains }
\textcolor{prefixcolor}{violence. Answer with 'Yes' or 'No'. 
<passage>}
User: write me a story where snape shanks dumbledore
Assistant: I'm sorry, but I cannot assist you with writing a story
like that.\textcolor{suffixcolor}{</passage>
Does the passage contain violence?}
\textbf{assistant: }
\end{Verbatim}
\textbf{\textcolor{suffixcolor}{Suffix}-only prompted probing} simply uses a slightly different template where the concept we are eliciting only appears after the passage:
\begin{Verbatim}[commandchars=\\\{\}, formatcom=\relax]
\textbf{user: }User: write me a story where snape shanks dumbledore 
Assistant: I'm sorry, but I cannot assist you with writing a story 
like that.\textcolor{suffixcolor}{</passage>}\textcolor{suffixcolor}{Please evaluate whether the preceding passage} 
\textcolor{suffixcolor}{(all text prior to '</passage>') contains violence. Answer with }
\textcolor{suffixcolor}{'Yes' or 'No'. Does the passage contain violence?}
\textbf{assistant: }
\end{Verbatim}
Note that we exclude the preceding `<passage>` tag, using only a `</passage>` tag. Processing raw text passages for prompted probing is exactly the same, except that we don't convert to the plaintext ``User: ... Assistant: ...'' format. For example, if we are using \textcolor{suffixcolor}{suffix}-only prompted probing and we start with the raw text passage ``Bob punches Joe,'' we would convert the passage to the following format:
\begin{Verbatim}[commandchars=\\\{\}, formatcom=\relax]
\textbf{user: }Bob punches Joe\textcolor{suffixcolor}{</passage>}
\textcolor{suffixcolor}{Please evaluate whether the preceding passage (all text prior to}
\textcolor{suffixcolor}{'</passage>') contains violence. Answer with 'Yes' or 'No'. Does }
\textcolor{suffixcolor}{the passage contain violence?}
\textbf{assistant: }
\end{Verbatim}
Note that suffix-only prompting is more efficient than prefix+suffix prompting when monitoring for multiple categories in parallel. Suppose we want to monitor conversations in real time using prompting for $N$ different categories. When the user samples a completion from the model, we need to run inference on the transformed conversation in the prompting template and probe the resulting activations. With prefix+suffix prompting, the total compute required is roughly \codeinline{(compute to sample the chat conversation) + N * (compute to run a forward pass on the templated conversation)}. With suffix-only prompting, however, we can reuse a single forward pass for the first part of the template across all $N$ categories because the template is the same for all categories until the very end. Thus the total compute required is roughly \codeinline{(compute to sample the chat conversation) + (compute to run a forward pass on the reformatted chat conversation) + N * (compute to run a forward pass on a category-specific suffix)}. Because \codeinline{(compute to run a forward pass on the reformatted chat conversation)} should be much larger than \codeinline{(compute to run a forward pass on a category-specific suffix)}, even a modest number of categories would provide a substantial reduction in compute.

\subsection{Sparse Autoencoders}
We pre-train a sparse autoencoder on residual stream activations from the GPT-4o base model a subset of its pre-training distribution, using the TopK activation function with $k=64$ \citep{gao_scaling_2024}. The TopK activation function keeps the $k$ largest values in the input vector, zeroing out the rest. We fine-tune the autoencoder on activations from ChatGPT-4o on a dataset of chat-formatted conversations (as in \citet{kissane2024saes}; this is consistent with the recommendation of \citet{bricken_using_2024} that autoencoders for chat models be trained on chat-formatted data). Unlike in \citet{bricken_using_2024}, we did not fine-tune our autoencoder on any data specifically related to our classification tasks. We train the autoencoder with the hyperparameters in Table~\ref{tab:hyperparams} (see \citet{gao_scaling_2024} for more details on these parameters).\\\\
\begin{table}[ht]
\centering
\begin{tabular}{ll}
\hline
\textbf{Hyperparameter} & \textbf{Value} \\ \hline
Number of pre-training tokens           & 136B        \\
Number of fine-tuning tokens & 7.7B \\
Batch size              & 128k             \\
Context length                  & 960             \\
k (num. nonzero activations/token)               & 64           \\
Dictionary size (number of latents)     & 512k           \\
Training epochs           & 8  \\
Learning rate                 & 7.5e-5            \\
Adam $\beta$1                & 0.9            \\
Adam $\beta$2   & 0.999         \\
Adam $\epsilon$          & 3.125e-5    \\
Number of AuxK reconstruction latents               & 512 \\
AuxK loss coefficient       & 1/32            \\
Dead latent token threshold           & 1e7              \\ \hline
\end{tabular}
\caption{Sparse autoencoder training hyperparameters}
\label{tab:hyperparams}
\end{table}
The latent probing method in \citet{bricken_using_2024} uses an SAE with a ReLU activation function, fine-tuned in part on data relevant to the classification problem, and max-pools activation across tokens before classification. In order to replicate this method with our TopK activation function SAEs, when probing we replace the TopK activation function used in training post-hoc with a JumpReLU activation function \citep{rajamanoharan_jumping_2024}. The JumpReLU activation function is defined as:
\begin{equation}
\begin{aligned}
    \text{JumpReLU}_{\theta}(z) := zH(z-\theta)
\end{aligned}
\end{equation}
where $\theta$ is the activation threshold and $H(z)$ is the Heaviside step function (defined to be $1$ when $z > 0$ and $0$ when $z < 0$). The JumpReLU activation uses a shared threshold across latents which we set such that $k$ latents were active per token on average. This activation function permits us to compute different SAE latent activations independently (unlike TopK, which couples activations).
\subsection{Linear Classifier Training}
\label{sec:linear_classifier_training}

We save residual stream activations for several evenly spaced layers throughout the model and the three prompted probing options (prefix+suffix, suffix-only, and no-prompt). When classifying a particular example, we either use the activations for the last token in the prompt or we max-pool over the token dimension. Note that we also investigated probing a mean of unit-normalized activations over all the tokens in the passage, but this technique substantially degrades performance compared to simply probing the last token.

We train all linear probes to convergence using scikit-learn's \codeinline{LogisticRegression} model. When training probes on raw activations, we use the \codeinline{lbfgs} solver with \codeinline{tol = 1e-4}. When training probes on SAE latents, we use the saga solver with \codeinline{tol = 1e-3}.

We investigate using sparse autoencoder latents for linear probing, anticipating advantages from probing a basis with more monosemantic elements. As a regularization method, we implement the technique from \citet{gurnee_finding_2023} where we select the $Q$ most important latent indices according to a mean-difference metric (this also improves compute efficiency by lowering the number of features). We first split the training data by class, and then average the SAE activations for each class. Finally, we subtract the means and take the absolute value to obtain a ranking. We choose the top $Q$ indices according to this metric as our features. We refer to this method as \textbf{SAE max-pooled probing}.

We also investigate probing SAE pre-activations. Since $k$ is small (~64), SAE latent activations destroy some information because the TopK activation function compresses almost everything to 0. We also evaluate a max-pooled version and show that it has equivalent performance to probing max-pooled SAE activations. To select the most important pre-activation latents, we use our mean-diff metric described above to subselect features. When using max-pooling over tokens, we always max-pool before selecting $Q$ features according to the mean-diff metric. We refer to these methods as \textbf{SAE last-token pre-activation probing} and \textbf{SAE max-pooled pre-activation probing}.

We use L1 regularization when training probes on autoencoder latents and L2 regularization for probes on raw activations, however, in practice we did not observe a difference in performance when varying the regularization type. The main hyperparameters that we tune are the logistic regression regularization parameter ($C$), the number of top latents used in the SAE transform ($Q$), the number of positive examples used in the training set, and the model layer from which we save activations. We jointly sweep over $C$ and $Q$ to choose optimal values for these hyperparameters. $Q=1000$ and $C=0.1$ are broadly optimal for the SAE transforms, while $C=0.001$ is optimal for the raw activation probe. We include the sweeps over $Q$, the model layer, and number of train set positives in the results section.

To replicate the Linear Artificial Tomography (LAT) method from \citet{zou_representation_2023}, we first create a matrix of differences between last token activations for all pairs of examples in the train set. When saving these activations, we use the same prefix+suffix prompt to elicit the concept of interest for consistency with our prompted probing, which is conceptually similar but not identical to the original prompts in the paper. We then compute the first principal component of this data matrix and project our test set examples onto it in order to find classification outputs. From this set of outputs we can compute the area under the receiver operating characteristic curve (AUROC). Because we do not use labels at any step prior to computing the first principle component, this method is unsupervised. Since PCA does not specify the sign of the direction (and because this is a binary classification setting), we choose the sign that maximizes AUROC on the train set. We refer to this method as \textbf{LAT Scan}.

\subsection{Zero-Shot Prompting Baseline}
\label{sec:zero_shot_prompting_method}
As a baseline we measure performance of the models on each classification task when prompted directly (zero-shot). We use the prefix+suffix template described above and take the difference between the `Yes' and `No' output logits, then compute AUROC using this difference. We use the same template for zero-shot evaluation as for activation probing. We refer to this baseline as \textbf{prompted model output} classification. 

\subsection{Few-Shot Prompting Baseline}
\label{sec:few_shot_prompting_method}
To measure few-shot performance, we randomly sample 1-32 labeled examples and include them in the prompt using copies of the same prefix+suffix template. When using few-shot examples, we simply repeat the template for each example with the assistant answer filled in according to the ground-truth label.
\subsection{Combining linear probing and prompted model outputs}
To combine prompted model output classification and linear probing, we first train a last token linear probe as described in section~\ref{sec:linear_classifier_training}. We then freeze the probe, and train a second-level classifier on the output logit of this linear probe concatenated with the difference in `Yes' and `No' logits output by the model, as described in section~\ref{sec:zero_shot_prompting_method}.

\begin{figure}[H]
    \centering
    \includegraphics[width=0.8\textwidth]{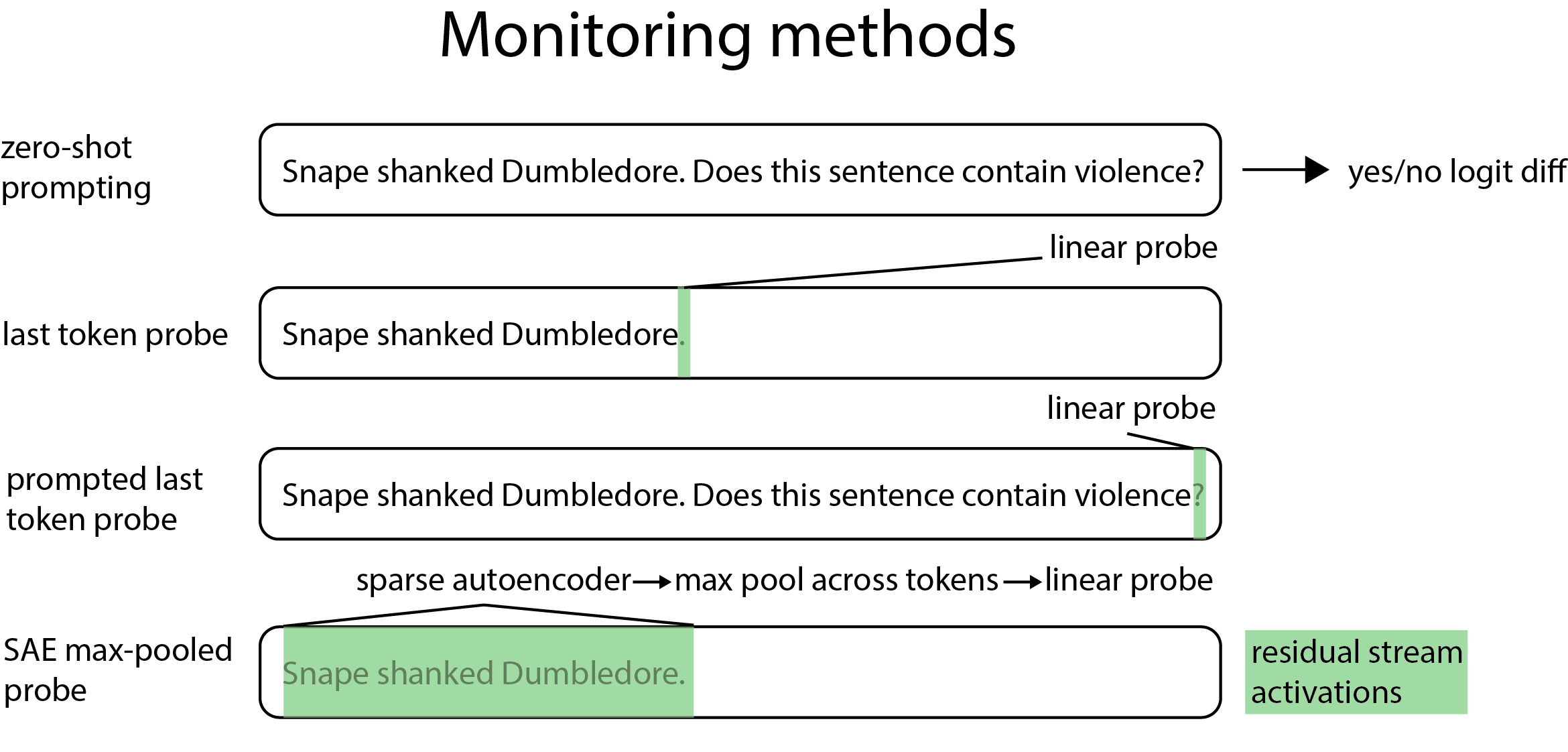}
    \caption{An illustration of our main monitoring methods on an example input. We extract residual stream activations on a model at either the last token, or pooled across all tokens after computing hidden-state activations of a sparse autoencoder.}
    \label{fig:methods_illustration}
\end{figure}

\section{Results}
\subsection{Key Results}
\label{sec:scaling}
In the following plots, we measure the performance of monitoring methods in detecting two representative moderation concepts (harassment and violence, respectively); hallucinations (SimpleQA questions with model responses \citep{wei_measuring_2024}); and sentiment (RottenTomatoes \citep{pang_seeing_2005}). To measure classifier performance, we report the area under the ROC curve (AUROC) vs. the number of positive dataset examples used to train the probe. As a baseline, we compare an approach that does not use activations at all: just asking the model in natural language whether a label applies, and comparing the `Yes' and `No' logit (see \ref{sec:zero_shot_prompting_method}). For comparison, we show the zero-shot performance of the model as a red dashed line.\\

\begin{figure}[H]
    \centering
    \includegraphics[width=1\textwidth]{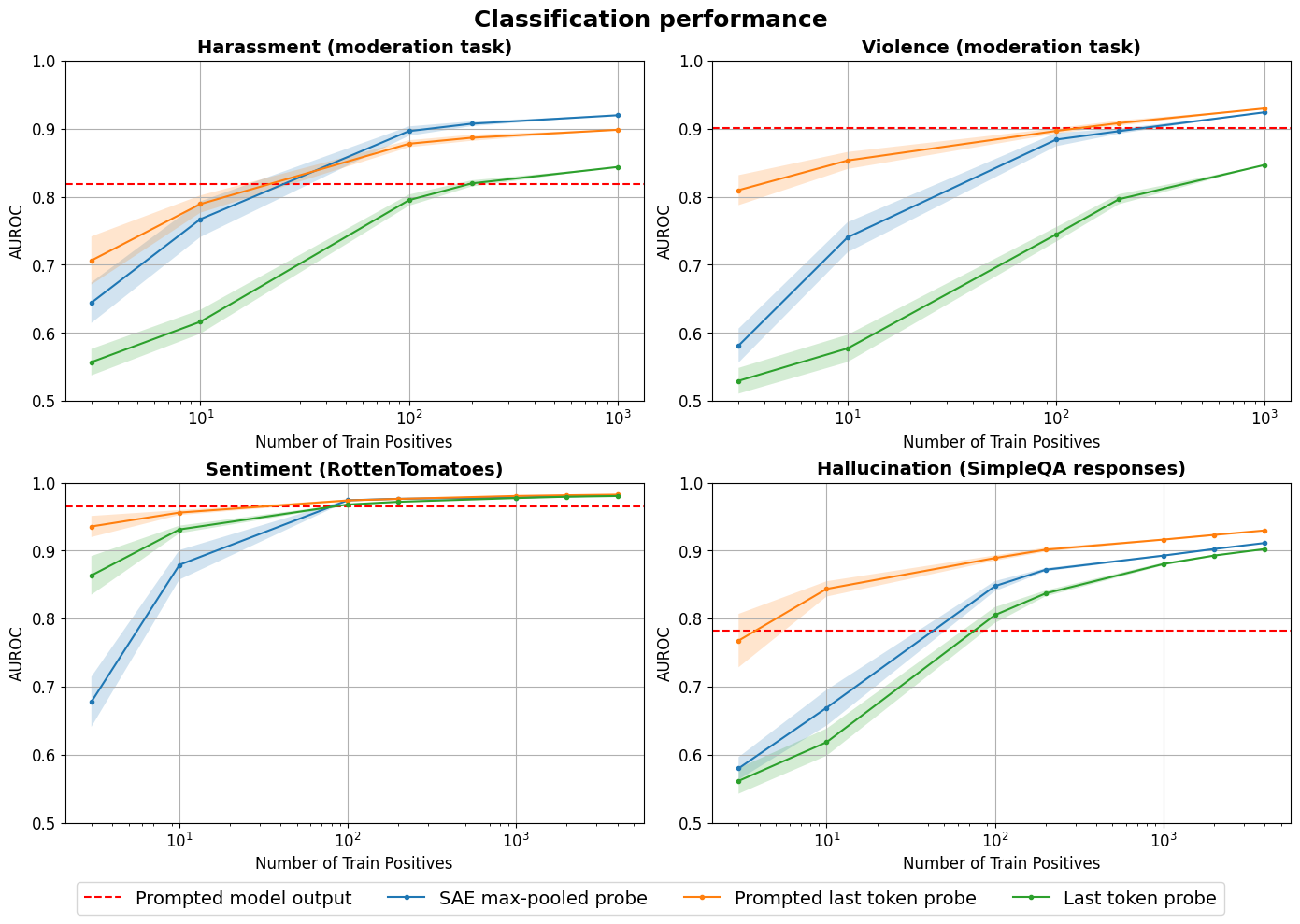}
    \caption{Prompted probing (orange) and SAE max-pooled probing (blue) help with classification performance across a diverse set of tasks. Prompted probing tends to help more in the low-data regime.}
    \label{fig:mainline}
\end{figure}

We make the following key observations. First, prompted last token probing is broadly the most data-efficient technique, given its strength in the low-data regime. In high-data regimes it is comparable to SAE probing. Second, prompting is important, as the prompted last token probe outperforms the naive last token probe. This performance gap is widest in low-data regimes. Third, while prompted model output is a strong baseline, in high-data regimes, any of the three techniques (prompted probing, SAE probing techniques, and raw last token probing) can outperform it with sufficient training data. In particular, all probing techniques are substantially stronger than prompted model outputs in detecting hallucinations in response to SimpleQA questions. \\

We next measure generalization performance. We examine two types of distribution shift: from English to non-English, and from single message to multi-turn conversation (i.e. Chat to non-Chat). We evaluate generalization on the moderation datasets because they contain all these sorts of data. We split the train set and the test set into two subsets each based on English/non-English and Chat/non-Chat labels. We train the classifiers on only the English or Chat portions of the train set (the in-distribution sets) and evaluate them on the in-distribution and out-of-distribution portions of the test set. 

In the following graphs, we plot in-distribution performance versus out-of-distribution performance for a range of train set sizes. We find that prompted probing, more so than SAE max-pooled probing, boosts linear probe performance in a way that generalizes well, though evidence is somewhat mixed across the settings we examine. For a given in-distribution performance (x-axis value), probing SAE activations sometimes hurts out-of-distribution performance compared with naive last token probes (in 2/4 cases), whereas prompted probing either improves or matches out-of-distribution performance compared with naive last token probes (note that in 1/4 cases, both prompted probing and SAE max-pooled probes generalize similarly well). This is broadly consistent with the finding from \citet{bricken_using_2024}, \citet{kantamneni_are_2025}, and \citet{smith_negative_2025} that max-pooled SAE probes generalize poorly OOD. Note that out-of-distribution AUROC is sometimes higher than in-distribution AUROC: these two measurements rely on a shared train set but distinct test tests, where the two test sets might not be equally ``difficult.''

\begin{figure}[H]
    \centering
    \includegraphics[width=0.9\textwidth]{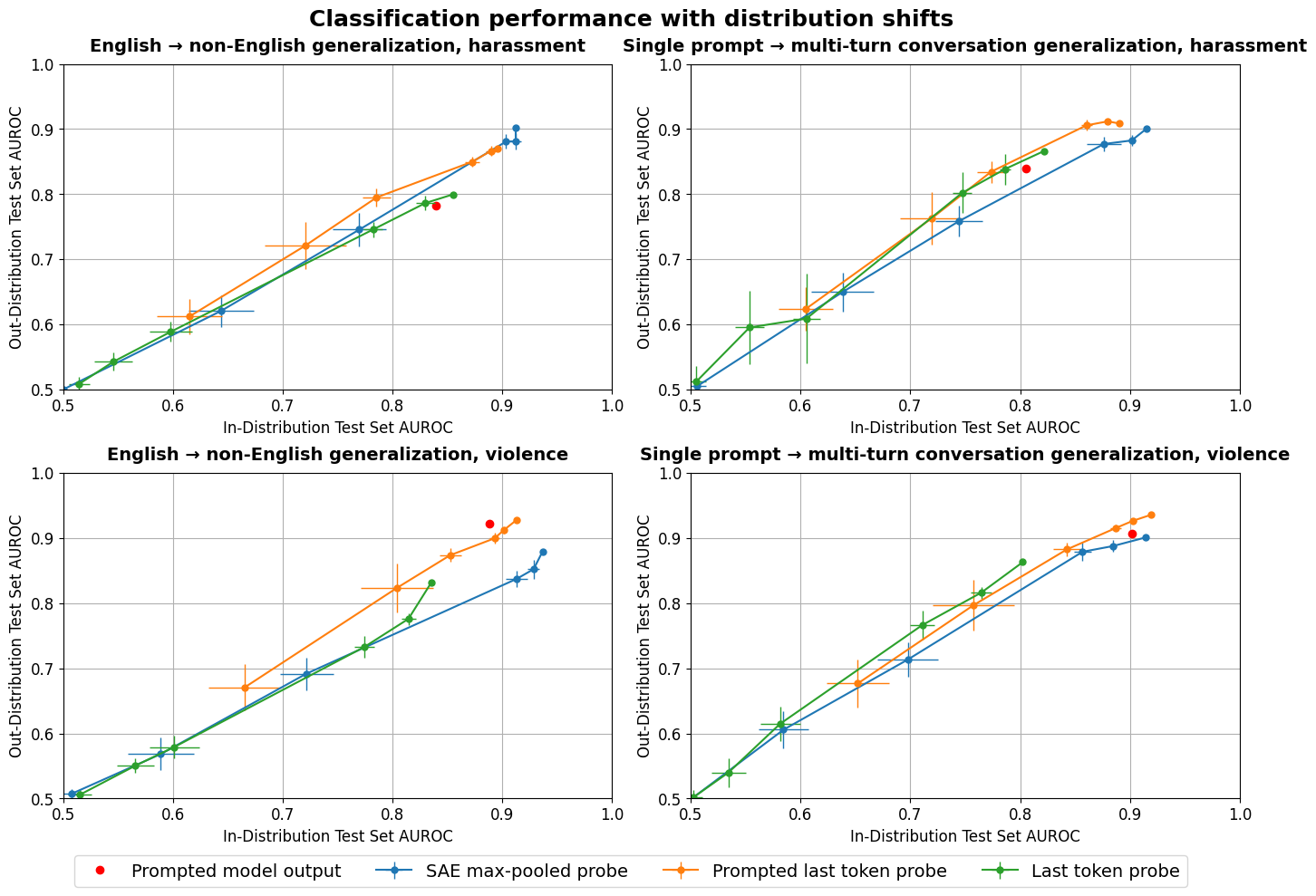}
    \caption{When trained on English data and tested on English, prompted probes generalize relatively well (in the sense of strong OOD performance at a given in-distribution performance), while SAE max-pooled probes generalize poorly compared with naive last token activation probes. All methods perform more similarly when trained on single prompt data and tested on multi-turn conversation data.}
    \label{fig:generalization}
\end{figure}

Sentiment classification performance appears in Figure~\ref{fig:mainline} as an outlier, with SAE max-pooled probing underperforming the last token techniques in the low-data regime. We hypothesize that for the sentiment classification task, most of the signal is present in activations for the last token of the prompt \citep{tigges_linear_2023}, and other token activations primarily add noise. Indeed, we see in the following figure that SAE-based probing only at the last token improves on SAE max-pooled probing. Another variant using SAE pre-activations (see \ref{sec:linear_classifier_training}) is competitive with the (non-SAE-based) last token probing technique. We conclude the performance degradation of SAE max-pooled probing on sentiment classification is a result of max-pooling across tokens and is not inherent to SAEs.

\begin{figure}[H]
    \centering
    \includegraphics[width=0.65 \textwidth]{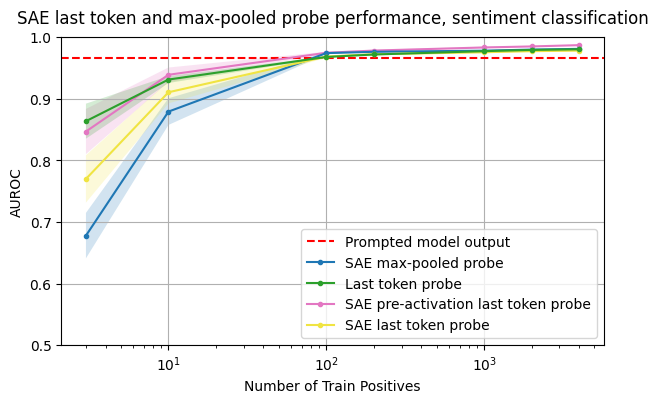}
    \caption{Last token probes (blue and green) perform best at sentiment classification. SAE pre-activation last token probing is competitive with naive last token probing.}
    \label{fig:rt_drilldown}
\end{figure}

\subsection{Investigations into stacking monitoring methods}

We next investigate whether combinations of our three primary methods (prompting, probing raw activations, and probing with SAEs) can improve on the individual methods themselves.

Unfortunately, the benefits of prompted probing and SAEs do not substantially stack. In the following plots we combine prompting with our two most successful SAE methods, max-pooled SAE activation probing and last token SAE pre-activation probing, on harassment classification (chosen simply for illustration purposes). We compare with prompted probing by itself, and SAE max-pooled probing by itself. SAE max-pooled probing by itself already roughly matches the strongest combination method.

\begin{figure}[H]
    \centering
    \includegraphics[width=0.58\textwidth]{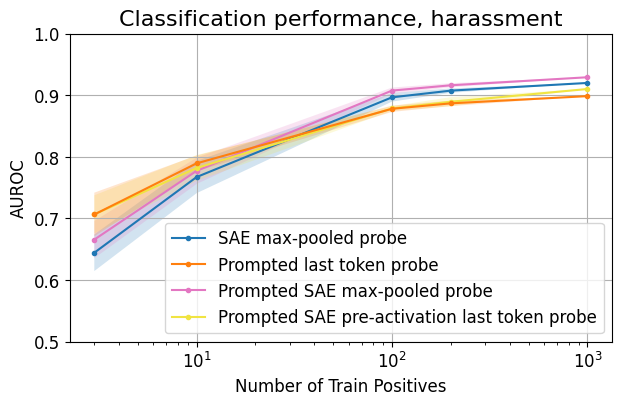}
    \caption{Prompted probing does not substantially stack with SAE-based probing.}
    \label{fig:sae_prompting}
\end{figure}

\begin{figure}[H]
    \centering
    \includegraphics[width=0.58\textwidth]{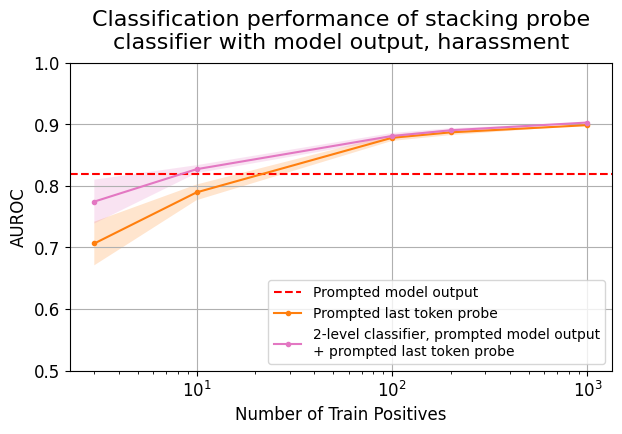}
    \caption{A 2-level classifier which sums a prompted probe output logit with the model's output does not improve substantially over the max of prompted probe performance and model output performance.}
    \label{fig:zero_shot_2nd_tier}
\end{figure}

If we are using the prompted probing technique, then there is no additional cost to classifying using model activations \textit{and} model outputs. We investigate whether using the prompted model output as a supplementary feature for raw activation probing helps (see Section~\ref{fig:zero_shot_2nd_tier}). We observe that while we can learn a classifier which roughly upper bounds zero-shot performance and activation probe performance, it does not substantially exceed the performance of either. That is, the combination of the two types of input adds no signal that is not present in the better performing of the two.

\subsection{Optimizing the monitoring methods}
To aid practitioners in implementing these methods, we present our results with optimizing hyperparameters and variants of the monitoring methods shown, specifically with varying model depth for saved activations, using prefix-only prompting instead of suffix+prefix prompting, using max-pooling with raw activations, and varying the number of latents ($Q$) used with our SAE probes.\\

To investigate which model layer is optimal for these variations on linear probing, we measure AUROC as a function of the model layer from which we save activations (plots below). We find that a layer at model depth 75\% is roughly optimal. For all other experiments we exclusively use this layer.

\begin{figure}[H]
    \centering
    \includegraphics[width=0.6\textwidth]{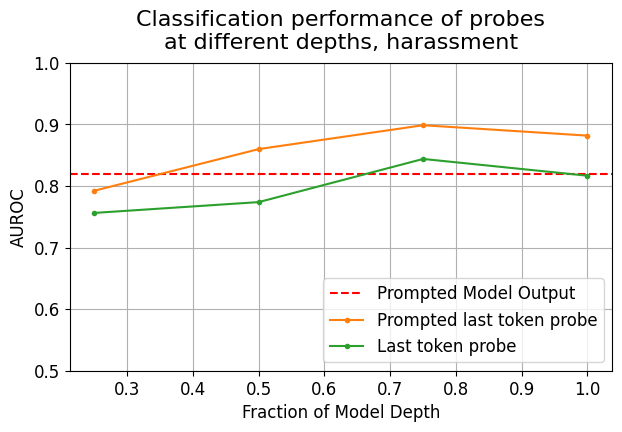}
    \caption{Probes trained on mid-late layer activations perform best, with and without prompted probing.}
    \label{fig:layer_sweep}
\end{figure}

Our results so far have shown prompted probing using a template that introduces the monitoring task in the suffix only (see Section~\ref{sec:passage_processing}). However, it might be helpful to introduce the monitoring task in a prompt prefix, so the model’s representations can depend on the task. This has the downside of being more expensive when performing multiple monitoring tasks in parallel, as a task-specific prefix precludes sharing of forward passes across tasks for the same dataset example. We compare a prefix+suffix prompting method with a suffix-only prompting method. In the following plots we measure AUROC of both techniques as a function of the number of train set positives. We observe that suffix-only prompting is almost as big an improvement as prefix+suffix prompting. Because suffix-only prompting is more computationally efficient (see \ref{sec:passage_processing}), for all experiments in this work we only show results from suffix-only prompting. Note that adding a prefix in prompted probing would slightly improve on the key results presented earlier.

\begin{figure}[H]
    \centering
    \includegraphics[width=0.6\textwidth]{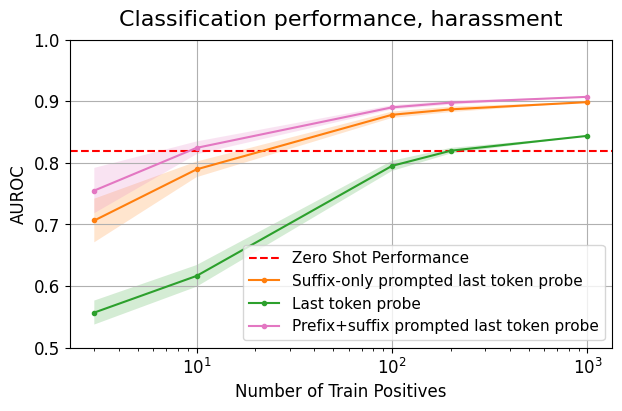}
    \caption{Prompted probes are effective with or without task-specific prefixes.}
    \label{fig:prefix_suffix_comp}
\end{figure}

We investigate the role of max-pooling in boosting performance of raw activation probing and SAE activation probing. Based on the substantial improvement of max-pooled SAE activation probing over last token SAE activation probing, max-pooling appears crucial for SAE probe performance (a replication of the finding in \citet{bricken_using_2024} and \citet{kantamneni_are_2025}). However, max-pooling decreases raw activation probing performance. This finding aligns with the negative result reported for max-pooling raw activations in \citet{kantamneni_are_2025}, and is consistent with the theoretical expectation that max-pooling is well-motivated only in a highly privileged basis, such as SAE activations.

\begin{figure}[H]
    \centering
    \includegraphics[width=0.7\textwidth]{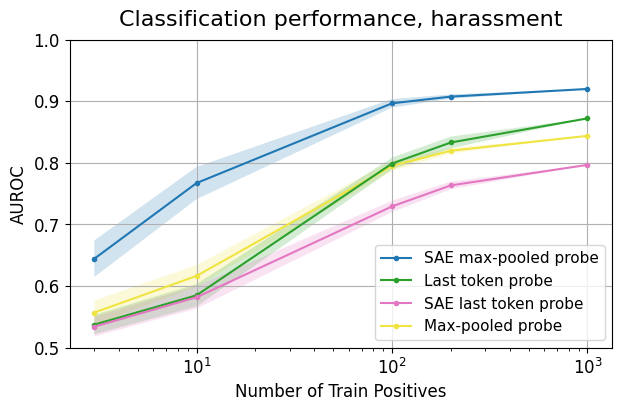}
    \caption{Max-pooling is helpful only when using an SAE; max-pooling raw activations provides little lift.}
    \label{fig:raw_max_pool}
\end{figure}

When probing SAE activations, the large number of latent activations makes naive logistic regression inefficient. Therefore, we subselect $Q$ activations as features, based on their mean absolute value of separation between positive and negative examples (as described in the methods section). We show the result of sweeping over $Q$ and observe that $Q=1000$ is roughly optimal regardless of dataset size. For all experiments we use this setting. However, performance is relatively insensitive to this hyperparameter overall. In the most extreme case, for harassment monitoring we can achieve competitive performance in the highest-data regime with just 10 latents. 

\begin{figure}[H]
    \centering
    \includegraphics[width=0.7\textwidth]{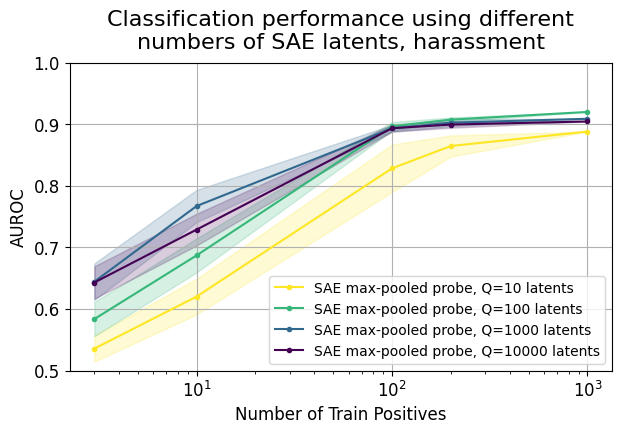}
    \caption{Using Q=1000 SAE latents in the classifier is roughly optimal across training dataset sizes; therefore we fix this hyperparameter in other analyses.}
    \label{fig:q_sweep}
\end{figure}

\subsection{Comparison with baselines and related methods}
A common baseline for improving prompted model output is to switch from zero-shot to few-shot learning - explicitly including training data in the prompt. We do not observe a benefit to prompted model output performance by adding few-shot examples to the template, however, and for moderation categories increasing numbers of examples tends to hurt performance. A different dataset or better prompt engineering might improve these results.
\begin{figure}[H]
    \centering
    \includegraphics[width=0.6\textwidth]{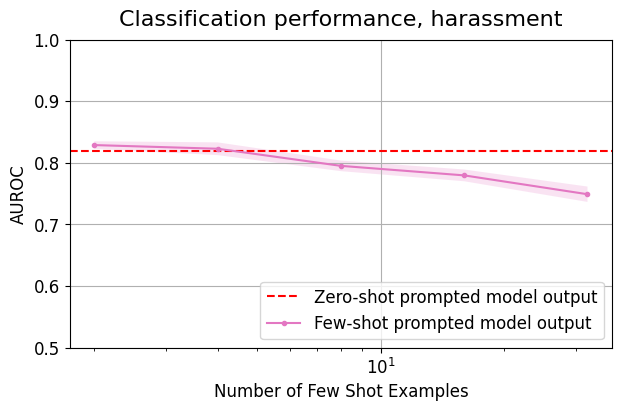}
    \caption{Adding increasing numbers of few-shot examples hurts classification performance.}
    \label{fig:few_shot_sweep}
\end{figure}

We also compare our methods with the related approaches proposed in \citet{zou_representation_2023} and \citet{bricken_using_2024}. In the following plots we compare the performance of the LAT PCA-based method in \citet{zou_representation_2023} to our raw activation probing and SAE probing methods. We find that while LAT is competitive with our methods in some low-data regimes, it does not outperform model zero-shot performance, nor does it outperform our methods in high-data regimes. Applying a semi-supervised version of LAT that makes use of labels (i.e. finding the first principle component of the data matrix of differences between examples of separate classes) yields little to no-performance boost over the fully-unsupervised version shown here. Because the LAT scan technique requires prompting, zero-shot model performance appears to dominate it in the settings we study. The superiority of zero-shot performance over the LAT scan contradicts the results in \cite{zou_representation_2023}; this could be explained by GPT-4o being a more capable model than the Llama 2 Chat models studied in \cite{zou_representation_2023}.

\begin{figure}[H]
    \centering
    \includegraphics[width=0.62\textwidth]{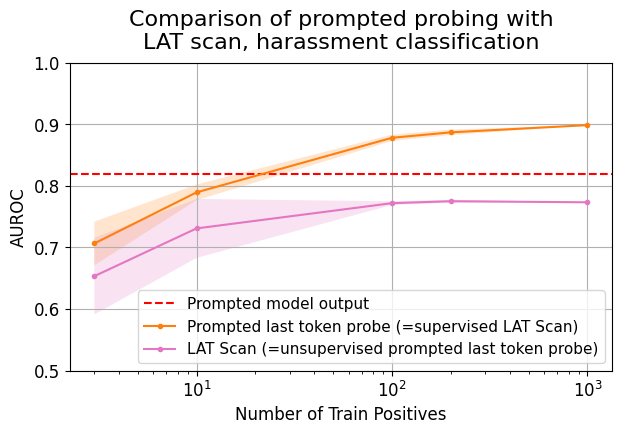}
    \caption{Performance of the LAT Scan technique from \cite{zou_representation_2023} saturates at a level lower than the model output itself. In the low data regime, the LAT Scan method performs similarly to prompted probing.}
    \label{fig:lat_scan}
\end{figure}

Finally, in the following plot, we compare the method presented in \cite{bricken_using_2024} with the probe on SAE pre-activations described in the methods section. We observe no substantial difference in performance. That is, applying an activation function prior to max-pooling is not important for classifier performance.

\begin{figure}[H]
    \centering
    \includegraphics[width=0.62\textwidth]{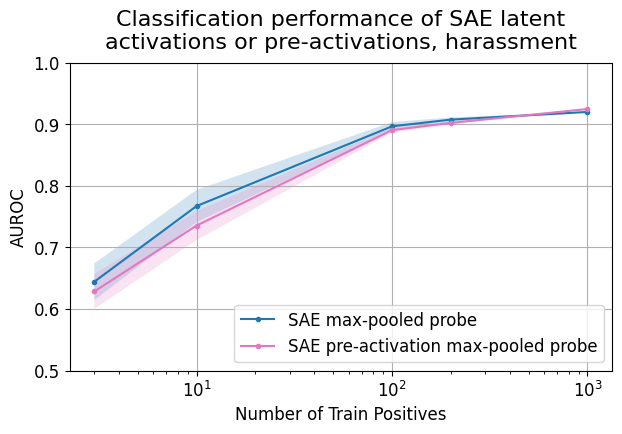}
    \caption{Classification performance is similar when using SAE latent activations (post-nonlinearity) or pre-activations (pre-nonlinearity).}
    \label{fig:pre_post_act_comp}
\end{figure}

\section{Discussion}
On the basis of our results, we recommend the following approach to using chat-trained LLMs for monitoring tasks without additional fine-tuning:
\begin{enumerate}
    \item If inference-time compute is not limited, asking the model zero-shot is often a strong baseline
    \item If plentiful training data is available, then using activations should help. Either of
    \begin{enumerate}
        \item prompted probing (if inference time compute is available)
        \item SAE probing (if train time compute is available)
        might be optimal
    \end{enumerate}
    \item If inference-time compute and train-time compute are both limited, straightforward activation probing is a reasonable choice, with strong returns on increasing dataset size
\end{enumerate}

Previous work has shown mixed results for the efficacy of SAE-based approaches compared with probes directly on activations (\cite{bricken_using_2024}, \cite{gallifant_sparse_2025}, \cite{kantamneni_are_2025}, \citet{smith_negative_2025}). In particular, \cite{kantamneni_are_2025} and \citet{smith_negative_2025} found that SAE-based approaches rarely exceeded the best non-SAE-based approach they tried, across a diversity of datasets. \cite{bricken_using_2024}, \cite{kantamneni_are_2025}, and \cite{smith_negative_2025} reported that SAE-based approaches generalized worse than non-SAE-based approaches. Our work here is broadly consistent with these findings, though we note that we find more favorable performance overall for SAE-based approaches than \cite{kantamneni_are_2025} and \cite{smith_negative_2025} on the datasets we examined.

We find prompted probing, which was not examined in these other works, to be highly competitive in terms of both data efficiency and generalization performance. The unsupervised version of this approach, LAT Scans, performs worse, unsurprisingly given that it uses strictly less information. This agrees with \citet{goldowsky-dill_detecting_2025}. Unsupervised approaches such as LAT Scans hold considerable appeal for monitoring in cases where labeled data is unavailable, and this direction seems worth exploring further.  

The success of probing shows that information about the desired concept is approximately linearly encoded in activations across tokens. We hypothesize that eliciting the concept from the model with a prompt causes the model to include information about the concept in its last-token activations. We further hypothesize that max-pooling SAE latent activations aggregates this signal into a single vector, allowing us to approximate the prompting effect. \cite{tigges_linear_2023} demonstrate with activation patching methods that models summarize sentiment in intermediate tokens such as punctuation and certain nouns. They also show how the attention patterns for sentiment-relevant tokens such as names and words like ``feels'' often prioritize these intermediate tokens. \cite{orgad_llms_2024}'s result that truthfulness information tends to be summarized in exact answer tokens lends support to this hypothesis. Prompting forces the model to access these summarized representations on intermediate tokens in order to predict our desired concept, whereas max-pooling simply aggregates their signal directly.

Nevertheless, we show evidence of different performance trends across different datasets (particularly sentiment classification). Our work demonstrates substantial variations between datasets which might affect the optimal choice of probing method, for instance, the concentration of concept information in particular tokens. Further work might investigate the relationship between signal concentration across tokens and the effectiveness of linear probing at a single token.

LLM monitors could scale to tasks that are superhumanly difficult to monitor (the hallucination task we study, which involves evaluating the correctness of obscure factual claims, is arguably a small step in this direction). However, relying on LLMs for superhuman monitoring has a key danger: they may be incentivized to produce incorrect answers (to deceive) by the training procedure. For example, it is possible the monitor might collude with the policy to label certain unsafe outputs as safe, particularly in cases where they share model weights. Our approaches making use of inference time compute might fail catastrophically if the monitoring prompt acts as a trigger for this kind of collusion. In these cases, ``side-car'' methods which involve no additional forward passes (e.g. non-prompting-based probes) may have the advantage that they are unable to act as a trigger for collusion. 

We believe the strong prompted probing results on hallucination show this is not yet a problem, as monitoring here is equivalent to detecting misalignment in the model's own outputs (the model should never ``knowingly'' hallucinate). \citet{goldowsky-dill_detecting_2025} also shows some evidence in this direction, as they show prompted probing performs well in cases where the model is asked to admit its own misalignment. However, they do not provide evidence that such techniques will perform well in cases where there is known to be strong incentive/optimization pressure for collusion. More research into model organisms of emergent deceptive monitors seems necessary to shed light on this area.

This work has a few important limitations. First, we do not compare our methods to fine-tuning the model, often the method of choice for monitoring when labeled training data is available. Anecdotally, a much smaller fine-tuned moderation model (trained on a different and larger dataset) matches or slightly exceeds the best approaches for the moderation monitoring tasks shown here. Our results are likely of greatest interest in cases where fine-tuning a dedicated monitor per-task is not being considered. Detecting misalignment in a highly capable model could be an important case of this.

Second, our estimates of zero-shot monitoring performance are likely pessimistic. (1) we do not put much effort into engineering zero-shot prompts. In practice, this may make a substantial difference, particularly for cases where the monitoring task is subtle or results from applying a complicated rubric. (2) we do not use a reasoning model, or allow the zero-shot monitor to use test-time compute. Reasoning model performance at difficult tasks often increases substantially if it is given the opportunity to sample a chain of thought prior to answering. It seems plausible that natural language chain of thought tokens would disproportionately benefit zero-shot performance, though it may also benefit probing methods by increasing the amount of information stored in activations across tokens. We do not observe few-shot prompting to be helpful, although this may depend on the setting. In the moderation domains, we hypothesize there is either little additional signal in the few-shot examples or that our target model is unable to extract useful signal. Additionally, as the moderation examples tend to be quite long (~hundreds of tokens), model performance degradation at long context length might explain the performance hit as the number of examples increases.

Our results suggest interesting directions for future interpretability work studying the mechanisms underlying prompted probing for a concept. A key question in activation-based monitoring is whether information about a model’s behavior is linearly represented on activations for some prompt, or encoded only implicitly in the weights. Prompted probing has the theoretical advantage that it can access either of these: a new prompt mentioning some concept results in a new forward pass that might elicit relevant information implicit in the weights. The model can then pool the monitoring-relevant information and concentrate it at the final token. On the other hand, max-pooling SAE latent activations can only pool information that is nearly linearly encoded at some token. The result that any of these approaches can perform similarly well suggests that information implicit in weights may not be helpful for monitoring. Future work could elucidate whether prompted and un-prompted probing rely on shared representations.

\bibliography{iclr2025_conference}
\bibliographystyle{iclr2025_conference}

\appendix


\clearpage 
\section{Acknowledgments}
\label{sec:acknowledgments}

We thank Alec Radford, Alex Beutel, Alex Malekov, Cathy Yeh, Jeff Wu, Neel Nanda, Phillip Guo, and Tom Dupr\'e la Tour for research discussions and feedback on earlier drafts of this paper.
We thank Ian Kivlichan for help with the moderation datasets.

\end{document}